%% file: acl_latex.tex
\pdfoutput=1

\documentclass[11pt]{article}

\usepackage[preprint]{acl}
\usepackage{subfigure}

\usepackage{times}
\usepackage{latexsym}

\usepackage[T1]{fontenc}

\usepackage[utf8]{inputenc}

\usepackage{microtype}

\usepackage{inconsolata}

\usepackage{graphicx}
\usepackage{makecell}
\usepackage{booktabs}
\usepackage{times}
\usepackage{latexsym}
\usepackage{float}
\usepackage{graphicx}
\usepackage{amsmath}
\usepackage{multirow}
\usepackage{booktabs}
\usepackage{rotating}
\usepackage{subcaption}
\usepackage{tcolorbox}
\usepackage{threeparttable}
\usepackage{pythonhighlight}
\usepackage{listings}
\usepackage{algorithm}
\usepackage{algorithmic}
\usepackage{tabularx}
\usepackage{makecell}
\usepackage{enumitem}
\usepackage{adjustbox}
\usepackage{graphicx}
\usepackage{longtable}
\usepackage{colortbl}

\newcommand{\babyemoji}{\includegraphics[height=2\fontcharht\font`\B]{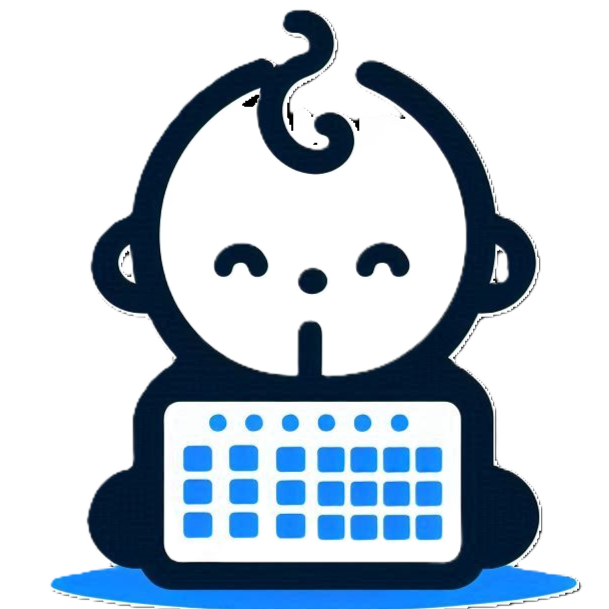}}

\title{\babyemoji{} \textit{ScratchEval: Are GPT-4o Smarter than My Child?}\\Evaluating Large Multimodal Models with Visual Programming Challenges}

\author{%
Rao Fu$^{\heartsuit}$,~
Ziyang Luo$^{\heartsuit}$,~
Hongzhan Lin$^{\heartsuit}$,~
Zhen Ye$^{\clubsuit}$,~
Jing Ma$^{\heartsuit}$\thanks{Corresponding Authors.}~
\\[4pt]
$^\heartsuit$Hong Kong Baptist University, $^\clubsuit$Hong Kong University of Science and Technology\\[3pt]
 \texttt{majing@comp.hkbu.edu.hk}
}

\begin{document}
\maketitle

\input{0.abstract}
\input{1.introduction}
\input{3.benchmark}
\input{4.results}

\input{5.conclusion}
\input{5.5reflection}

\bibliography{custom}

\appendix
\newpage

\input{6.appendix}



\end{document}

%% file: 0.abstract.tex
\begin{abstract}
     Recent advancements in large multimodal models (LMMs) have showcased impressive code generation capabilities, primarily evaluated through image-to-code benchmarks. However, these benchmarks are limited to specific visual programming scenarios where the logic reasoning and the multimodal understanding capacities are split apart. To fill this gap, we propose \textbf{ScratchEval}, a novel benchmark designed to evaluate the visual programming reasoning ability of LMMs. \textbf{ScratchEval} is based on Scratch, a block-based visual programming language widely used in children's programming education. By integrating visual elements and embedded programming logic, \textbf{ScratchEval} requires the model to process both visual information and code structure, thereby comprehensively evaluating its programming intent understanding ability. Our evaluation approach goes beyond the traditional image-to-code mapping and focuses on unified logical thinking and problem-solving abilities, providing a more comprehensive and challenging framework for evaluating the visual programming ability of LMMs. \textbf{ScratchEval} not only fills the gap in existing evaluation methods, but also provides new insights for the future development of LMMs in the field of visual programming. Our benchmark can be accessed at \url{https://github.com/HKBUNLP/ScratchEval}.
\end{abstract}

%% file: 1.introduction.tex
\begin{figure}
    \centering
    \scalebox{1.0}{\includegraphics[width=\linewidth]{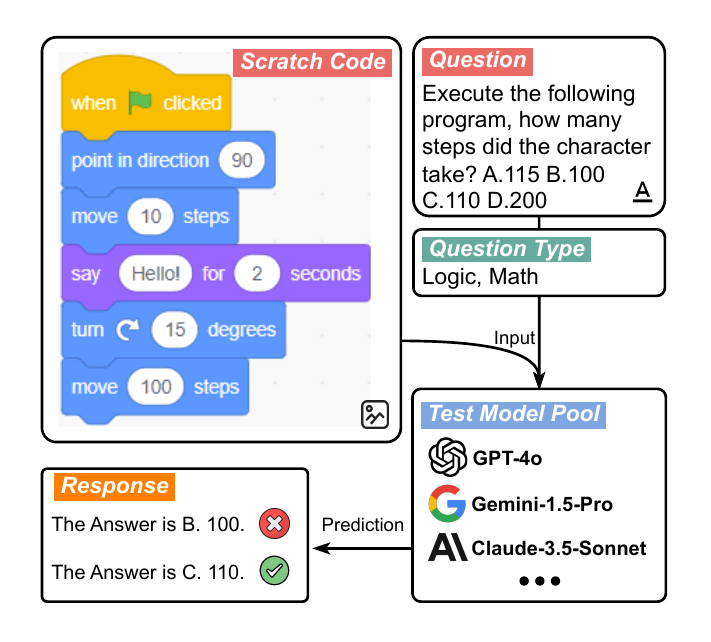}}
    \caption{The illustration of the evaluation process for \textbf{ScratchEval}.}
    \label{fig:task_illustration}
\end{figure}

\section{Introduction}

Recently, Large Multimodal Models (LMMs) such as GPT-4o~\cite{GPT4}, Gemini~\cite{gemini}, and Claude~\cite{claude} have shown remarkable capabilities in multimodal understanding~\cite{MMStar,lin2024goatbenchsafetyinsightslarge,wang2024mfcbenchbenchmarkingmultimodalfactchecking,luo2024videoautoarenaautomatedarenaevaluating,MM-Vet}. To assess their abilities, several comprehensive benchmarks have been introduced, including MMMU~\cite{MMMU}, MME~\cite{MME}, MathVista~\cite{MathVista}, and MMBench~\cite{MMBench}. These benchmarks primarily focus on evaluating core multimodal skills of LMMs, such as object detection, OCR, and visual reasoning. The evaluations provide deeper insights into the strengths and limitations of LMMs.

In addition to general multimodal understanding tasks, recent works such as MMCode~\cite{MMCode}, Design2Code~\cite{Design2Code},
Plot2Code~\cite{Plot2Code},
and CharMimic~\cite{ChartMimic} focus on assessing the visual programming reasoning abilities of LMMs. 
Most of the previous work focuses on specific scenarios, such as converting matplotlib images to Python code, generating code based on diagrams of algorithmic problems, or even generating HTML code from web page screenshots. Although these studies include visual elements, the diversity of input is relatively limited, mainly focusing on a single mapping from image to code, but ignoring cases where the programming logic is inherent in images.

In this paper, we argue that it is imperative to evaluate the visual programming capacity of  LMMs by unifying visual understanding and logical reasoning. Inspired by children's coding education~\cite{perez2020can}, using a graphical programming way, allows the assessment to focus more on logical thinking and problem-solving skills, rather than traditional programming languages that may be plagued by syntax errors. Thus we aim to combine visual elements with programming logic, requiring LMMs to process both visual information and code structure. 

To this end, we introduce \textbf{ScratchEval} as illustrated in Figure~\ref{fig:task_illustration}, a novel benchmark designed to assess LMMs' visual programming reasoning abilities by integrating visual elements with embedded programming logic. ScratchEval is based on Scratch~\cite{dasgupta2017scratch}, a popular block-based visual programming language widely used as an educational tool for children aged 8 to 16. It allows users to create projects through a drag-and-drop block interface, offering a visual approach to coding. By leveraging graphical code, our evaluation focuses on the complexity of multimodal input, where the model must understand the image, graphical programming language, and underlying logic, showcasing a comprehensive grasp of programming intent.

On ScratchEval, we tested multiple existing open-source and closed-source LMMs and studied the impact of different prompting strategies on model performance. Finally, we conducted a case study to analyze the performance bottleneck of the model. Through our research, we found that the existing state-of-the-art LMMs still fail to achieve high performance on our proposed benchmark, which shows the inadequacy of existing models in visual code reasoning capabilities and also points out the direction for further research.

%% file: 3.benchmark.tex
\section{ScratchEval}

All our data is manually collected and cleaned by experts from public question banks on the web. We organized the data into 305 multiple-choice questions, each with a problem description, options, and a picture containing the Scratch script and other necessary information.

Our test benchmark consists of two components: Chinese and English data. Both sections are identical in quantity and content, but the questions and Scratch script images are in their respective languages. This approach evaluates the visual reasoning capabilities of various models across different linguistic contexts, allowing us to assess how language-specific factors influence performance in interpreting visual information in Scratch programming. By comparing results from both datasets, we gain insights into the models' cross-linguistic robustness and adaptability.

\begin{table}[t]
  \centering
  \begin{tabular}{lc}
    \hline
    \textbf{Task} & \textbf{Number} \\
    \hline
    Math                    &  133          \\
    Logical thinking        &  99        \\
    Graphic perception      &  59          \\
    Spacial perception      &  43          \\
    \hline
    All                     &  305         \\
    \hline
  \end{tabular}
  \caption{Data volume of the four tasks, each question examines at most two types of the tasks.}
  \label{tab:datanumber}
\end{table}

\subsection{Data analysis}
Based on the content of the questions, we categorized them into four domains: mathematics, logical thinking, graphic perception, and spatial perception. The specific distribution of questions across these categories is presented in Table ~\ref{tab:datanumber}. It is important to note that some questions evaluate multiple abilities, and therefore, each question is assigned to at most two categories. The characteristics of each category are as follows:

\textbf{Mathematics tasks} encompass simple arithmetic problems typically encountered in elementary and junior high school curricula. These tasks assess the model's ability to solve basic mathematical problems.

\textbf{Logical thinking tasks} evaluate the model's capacity for logical reasoning based on provided Scratch scripts. These scripts are designed for children and are generally comprehensible even to those unfamiliar with the Scratch programming environment.

\textbf{Graphic perception tasks} examine the model's understanding of graphics. These may involve selecting graphics that correspond to a given script or inferring the output of a simple drawing program.

\textbf{Spatial perception tasks} assess the model's ability to determine the final position and orientation of a character based on a movement program.

This categorization enables thorough assessment of models' visual code reasoning abilities across cognitive domains.

\subsection{Evaluation Methodology}

The evaluation process consists of three stages: 1) generating answers, 2) extracting answers, and 3) calculating scores.

\begin{table*}[ht] \small
  \centering
  \renewcommand{\arraystretch}{1.2} 
  \setlength{\tabcolsep}{8pt} 
  \begin{tabular}{lcccccc}
    \toprule
    \textbf{Models} & \textbf{Size} &\textbf{All} & \textbf{Math} & \makecell{\textbf{Logical}\\ \textbf{Thinking}} & \makecell{\textbf{Graphic}\\\textbf{Perception}} & \makecell{\textbf{Spatial}\\\textbf{Perception}}\\
    \midrule
    \rowcolor{pink!50}
    \multicolumn{7}{c}{\textit{Proprietary Models}}\\
    \textbf{Gemini-1.5-Pro}      & -     & \textbf{52.8} & \textbf{55.3} & \textbf{49.5} & \textbf{47.5} & \textbf{59.5} \\
    \textbf{GPT-4o}     & -   & 43.9 & 44.7 & 42.4 & 45.8 & 50.0 \\
    \textbf{GPT-4-Turbo} & -  & 40.7 & 39.4 & 44.4 & 37.3 & 43.0 \\
    \textbf{Claude-3.5-Sonnet}    & -    & 40.3 & 45.5 & 37.3 & 35.6 & 35.7 \\

    \midrule
    \rowcolor{green!30}
    \multicolumn{7}{c}{\textit{Open-Source Models}}\\
    \textbf{Qwen2-VL}   & 72B      & \textbf{45.0} & \textbf{50.0} & \textbf{42.4} & \textbf{45.8} & \textbf{40.5} \\
    \textbf{LLaVA-v1.6} & 34B & 26.5 & 21.2 & 30.3 & 35.6 & 26.2 \\
    \textbf{InternVL2} & 26B & 22.3 & 25.6 & 18.2 & 20.3 & 21.4 \\
    \textbf{Pixtral} & 12B & 34.1 & 34.1 & 34.3 & 32.2 & 28.6 \\
    \textbf{MiniCPM-v2.6} & 8B & 30.0 & 28.0 & 31.3 & 39.0 & 31.0 \\
    \textbf{Molmo} & 7B & 31.2 & 32.6 & 29.3 & 33.9 & 26.2 \\
    
    \bottomrule
  \end{tabular}
  \caption{Accuracy (\%) of ten state-of-the-art LMMs on the English data of ScratchEval benchmark, tested across multiple cognitive abilities: math, logical thinking, graphic perception, and spatial perception.}
  \label{tab:experiment_results}
\end{table*}

First, the tested LMM generates answers based on the input query, which includes questions, options, image data, and a system prompt. After our experiments, the system prompt we set can help us greatly simplify the output of the model. Finally, the extracted answers are normalized to the required answer format option letters, and the target metric score is calculated. Using the fact that the examples in ScratchEval are multiple-choice questions with text answers, the accuracy score is used as a metric for deterministic evaluation.

%% file: 4.results.tex
\section{Experiments}

\subsection{Experiment setup}

We evaluate a total of 10 LMMs on ScratchEval under two setups: (a) Closed-source LMMs, including Gemini-1.5-Pro~\cite{reid2024gemini}, GPT-4-Turbo~\cite{achiam2023gpt}, GPT-4o, and Claude-3.5-Sonnet; (b) Open-source LMMs, including  Qwen2-VL~\cite{wang2024qwen2vlenhancingvisionlanguagemodels}, LLaVA-v1.6~\cite{liu2024llava}, InternVL2~\cite{chen2024internvl}, Pixtral~\cite{agrawal2024pixtral12b}, MiniCPM-v2.6~\cite{yao2024minicpm} and Molmo~\cite{deitke2024molmopixmoopenweights}. We use the accuracy as the evaluation metric. We provide implementation details in the Appendix \S\ref{sec:setup}.

\subsection{Experiment analysis}
We evaluated the performance of 10 state-of-the-art LMMs by drawing the practice of the LMSYS Chatbot Arena leaderboard on our proposed ScratchEval benchmark, incorporating both Chinese and English data. The 
experimental results on English data are presented in Table~\ref{tab:experiment_results}. To conduct a detailed analysis of the LMMs' capabilities, we categorized the questions into four domains: mathematics, logical thinking, graphic perception, and spatial perception.

The results reveal significant performance variations across models in each category, with most models surpassing the 25\% random guessing threshold. This indicates that LMMs possess some visual code reasoning capabilities, enabling them to process visual information alongside language comprehension.

Gemini-1.5-Pro demonstrated superior performance, achieving the highest scores across all categories. However, most other models struggled to exceed 50\% accuracy, highlighting current limitations in LMMs regarding visual code reasoning. We attribute this to a lack of high-quality visual-language paired data during training, as larger models like Gemini-1.5-pro and GPT-4o performed better. Additionally, the model's vision tokenizer may influence its visual reasoning capabilities.

Most models underperformed in mathematical and logical reasoning tasks, suggesting a deficiency in multi-step reasoning. Conversely, LMMs exhibited better performance in graphic and spatial perception tasks, demonstrating an understanding of concepts such as orientation and distance, which they can leverage for reasoning to some extent. The experimental results on Chinese data can be found in the Appendix \S\ref{sec:cndata}.

\subsection{Prompting strategies study}

We investigated the impact of prompt engineering on the visual code reasoning capabilities of models using our test benchmark. Previous studies, such as COT~\cite{wei2023chainofthoughtpromptingelicitsreasoning}, have shown that appropriate prompting can enhance the performance of large language models. However, its effectiveness for multimodal large language models remains underexplored. To address this, we selected four models and applied three prompting strategies to examine their influence on reasoning abilities.

The prompting strategies employed were: (1) Original prompt ("no-CoT"): using raw data as prompts. (2) zero-shot CoT ("CoT"): Chain of Thought prompting, appending "Let's think step by step." to each question for more comprehensive analysis. (3) eCoT: Inspired by ~\cite{ghosal2024languagemodelspuzzleprodigies}, we implemented eCoT, which requires a detailed examination during the CoT process by appending "Let's explain the picture and think step by step." to each question.

We found that CoT and eCoT techniques significantly enhanced the models' visual code reasoning capabilities, with CoT prompting improving performance by 10\% to 20\%. However, no model achieved overall accuracy exceeding 70\%, indicating substantial room for improvement. Additionally, eCoT yielded relatively minor improvements compared to CoT, suggesting that describing the image may hinder the model's visual code reasoning capabilities. Detailed experimental data can be found in the Appendix \S\ref{sec:allmodelsdata}

\begin{figure}[t]
    \centering
    \includegraphics[width=1\linewidth]{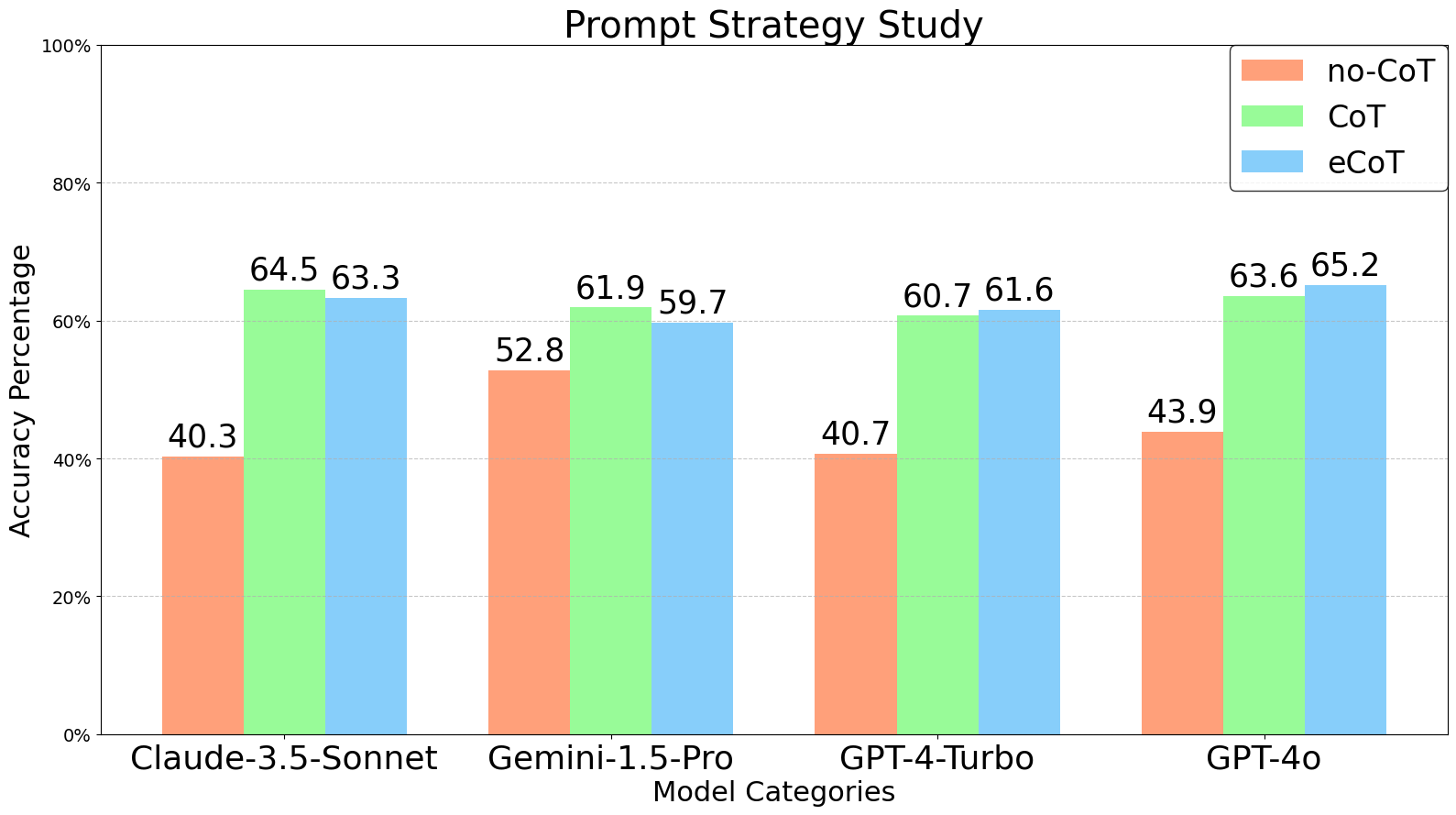}
    \caption{Models's performance under different prompting strategies.}
    \label{fig:allmymodels}
\end{figure}

\subsection{Case study}


To better understand the model's behavior, we selected several examples where Gemini-1.5-Pro made mistakes for a case study. Overall, Gemini-1.5-Pro is the best-performing model in ScratchEval. By studying its behavior, we aim to explain why ScratchEval is challenging for most models.

\begin{figure}[t]
    \centering
    \includegraphics[width=1\linewidth, height=0.32\textheight]{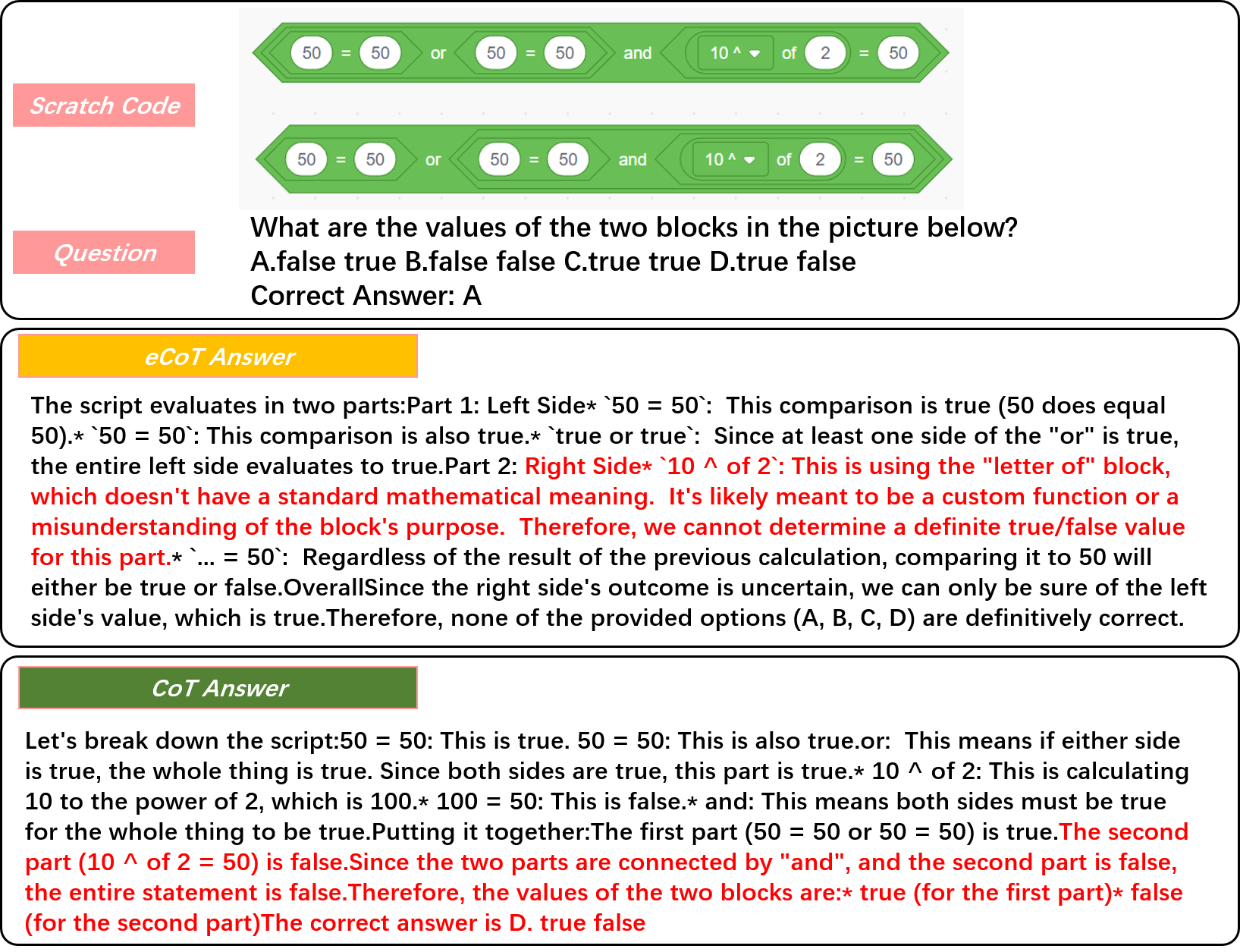}
    \caption{A Gemini-1.5-Pro mistake case. The error areas are marked in red.}
    \label{fig:eg0}
\end{figure}

We chose representative examples for Gemini-1.5-Pro's case study, as shown in Figure~\ref{fig:eg0}. We specifically selected examples that failed across all three prompting strategies mentioned earlier, allowing us to observe Gemini-1.5-Pro's deficiencies in certain areas.

As shown in Figure~\ref{fig:eg0}, Gemini-1.5-Pro with CoT accurately identified image content but hallucinated during reasoning. With eCoT, it described the image but misinterpreted symbols, leading to incorrect inferences.

These cases reveal that while Gemini-1.5-Pro excels in reasoning and basic math/logic problems, it struggles with subtle image distinctions. Visual encoders and hallucinations remain the main bottlenecks that restrict the model's reasoning capabilities. Additional examples are provided in Appendix~\ref{sec:twocase}.

%% file: 5.conclusion.tex
\section{Conclusion}

In this work, we present ScratchEval, a benchmark that uses the Scratch language to systematically evaluate the visual programming capabilities of state-of-the-art LMMs. Our evaluation of 10 representative LMMs indicates that while these models show some visual comprehension, they struggle with visual code reasoning. This highlights the need for research on models that integrate visual perception with logical thinking. ScratchEval provides a foundation for future studies aimed at enhancing AI systems' visual reasoning capabilities, bridging the gap between visual understanding and logical reasoning in LMMs.

%% file: 5.5reflection.tex
\section*{Limitations}

Although our proposed ScratchEval helps us to evaluate the visual reasoning ability of existing LMMs, we recognize that our work still has several important limitations: (1) Due to the difficulty of LMMs to directly operate graphical programming languages, in order to use graphical programming to examine the model's visual programming abilities, we model the problem as Multiple choice questions. (2) the narrow domain focus of our benchmark, concentrating solely on visual programming abilities, limits the generalizability of our findings. The results obtained cannot be extrapolated to assess other competencies of LMMs. These limitations underscore the need for continued research and development of more comprehensive evaluation methodologies for large multimodal models.

%% file: 6.appendix.tex
\section{Appendix}

\subsection{Experiments setup}
\label{sec:setup}

In our study, we conducted comprehensive evaluations of 10 state-of-the-art Large Multimodal Models (LMMs) on the ScratchEval benchmark. The following models were included in our experiments:

\begin{itemize}
    \item Gemini-1.5-Pro-Exp-0827
    \item GPT-4o-2024-05-13
    \item Claude 3.5 Sonnet
    \item GPT-4-Turbo-2024-04-09
    \item Qwen2-VL-72b-Instruct
    \item InternVL2-26b
    \item LLaVA-v1.6-34B
    \item MiniCPM-V 2\_6
    \item Pixtral-12b-2409
    \item Molmo-7B-D-0924
\end{itemize}

All models were evaluated using their respective latest versions available at the time of the experiment. To ensure consistency and reproducibility across all tests, we maintained a constant temperature setting of 0 for all models. This setting was chosen to produce deterministic outputs and facilitate direct comparisons between models.

For each model, depending on the task being performed, we use specific system prompts to explain the next task to the model.These system prompts are as follows:

\begin{itemize}
    \item For no-CoT tasks: "According to the displayed Scratch script and the given question, please choose a correct answer from the four options ABCD. You only need to find the correct option, and no analysis is required. "
    \item For CoT tasks: "According to the displayed Scratch script and the given question, please choose a correct answer from the four options ABCD. "
    \item For eCoT tasks: "According to the displayed Scratch script and the given question, please choose a correct answer from the four options ABCD. "
\end{itemize}

The system prompts when executing Chinese tasks are the translations of the above corresponding tasks.

\subsection{Chinese data experiments}
In Table~\ref{tab:cnexperiment_results}, We can see that the performance of most models is basically the same as in the English task, while some models perform better. We believe this is because some models use more Chinese data during training.

\subsection{Data example}
\begin{figure}[htbp]
    \centering
    \includegraphics[width=\linewidth]{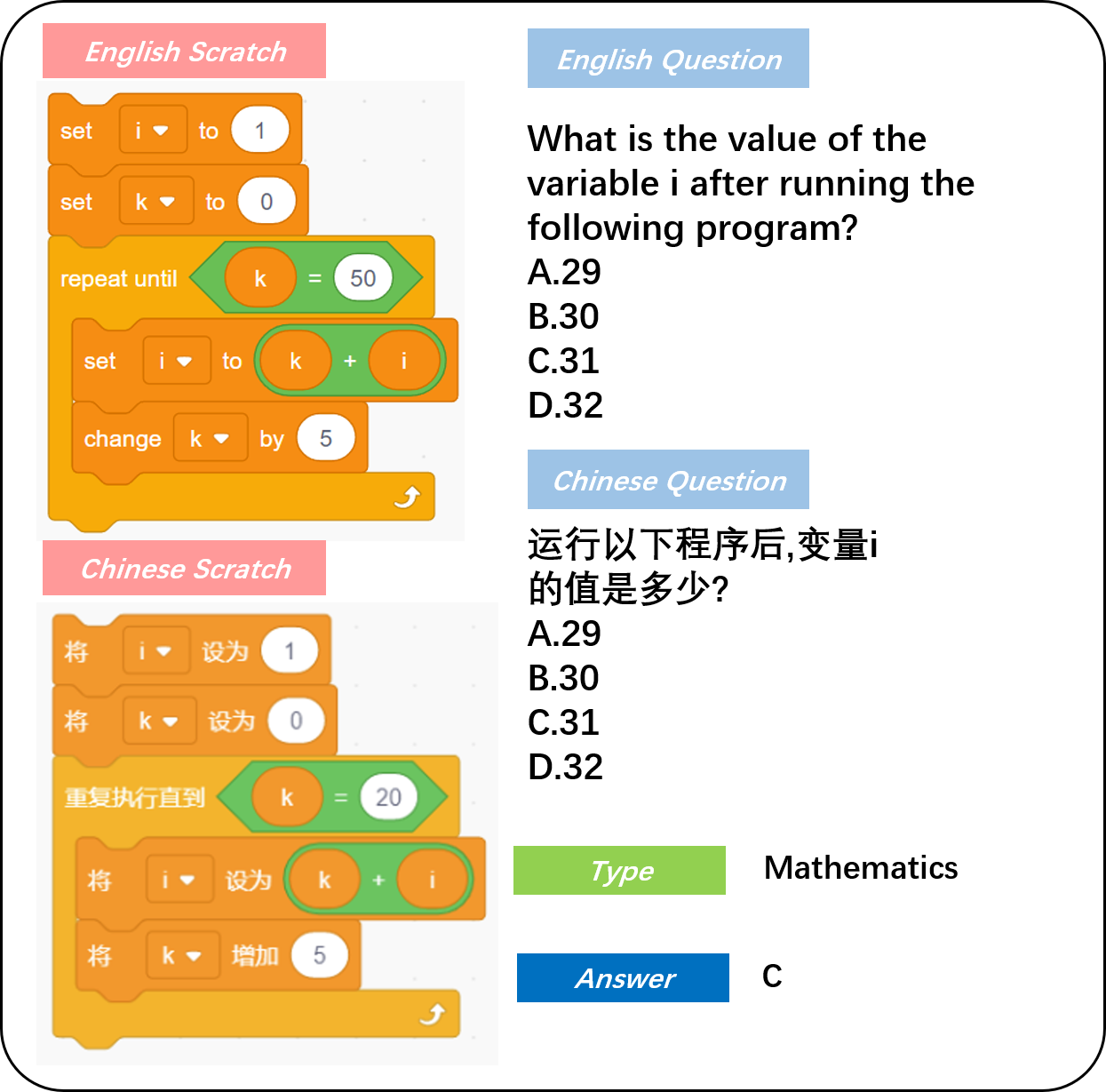}
    \caption{Data example about mathematics.}
    \label{fig:math}
\end{figure}

\begin{figure}[htbp]
    \centering
    \includegraphics[width=\linewidth]{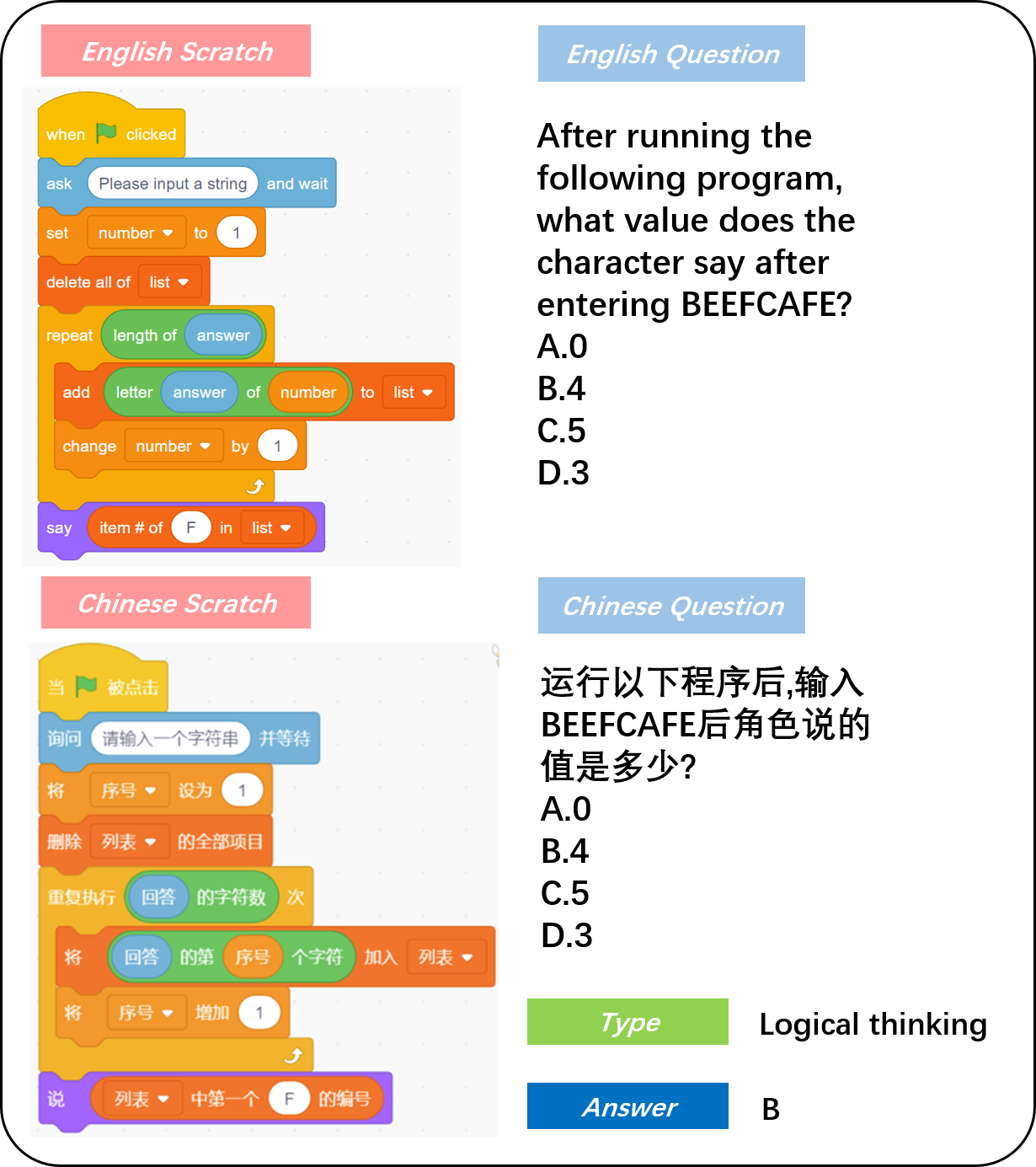}
    \caption{Data example about logic thinking.}
    \label{fig:logic}
\end{figure}

\begin{figure}[htbp]
    \centering
    \includegraphics[width=\linewidth]{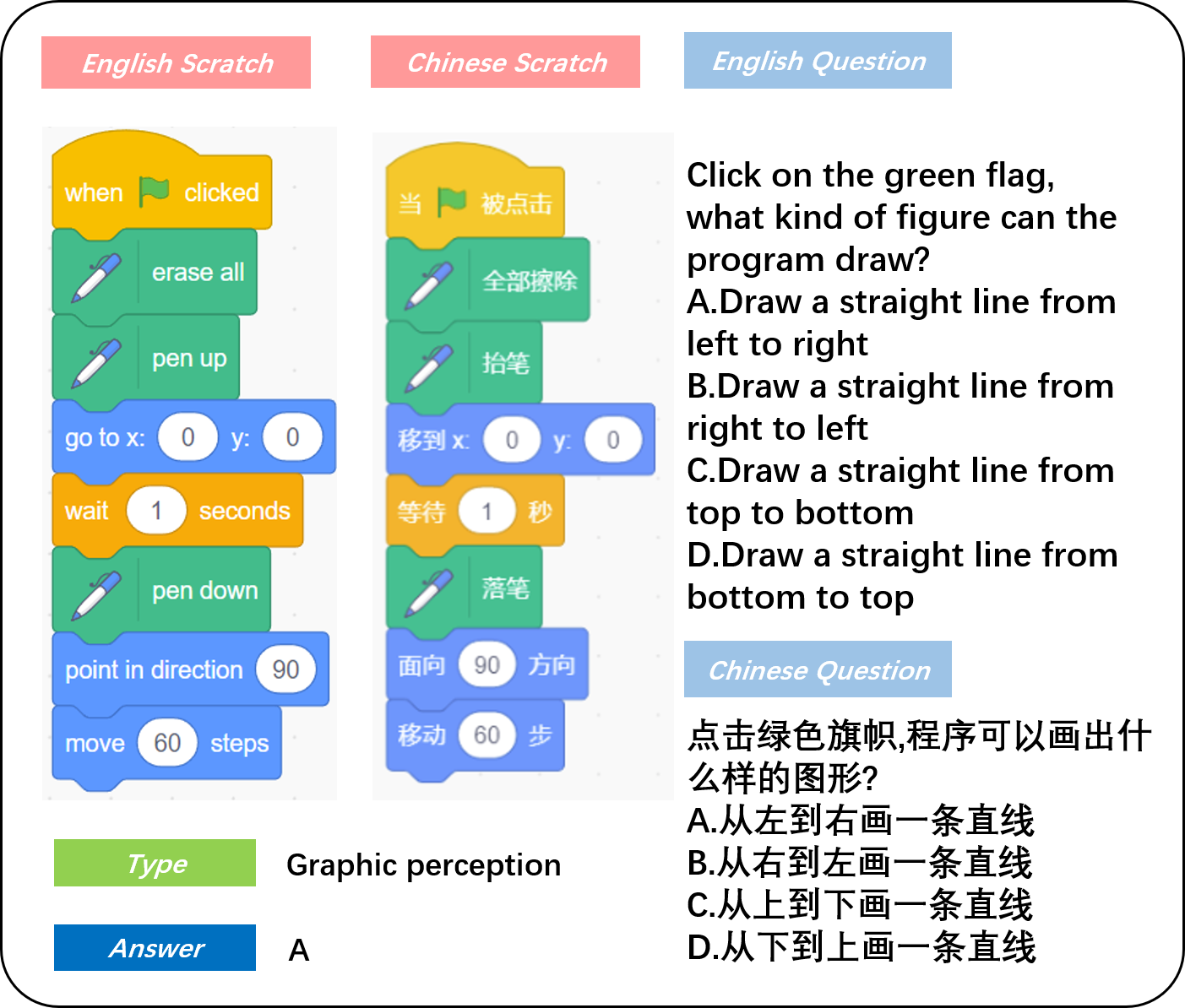}
    \caption{Data example about graphic perception.}
    \label{fig:graphy}
\end{figure}

\begin{figure}[htbp]
    \centering
    \includegraphics[width=\linewidth]{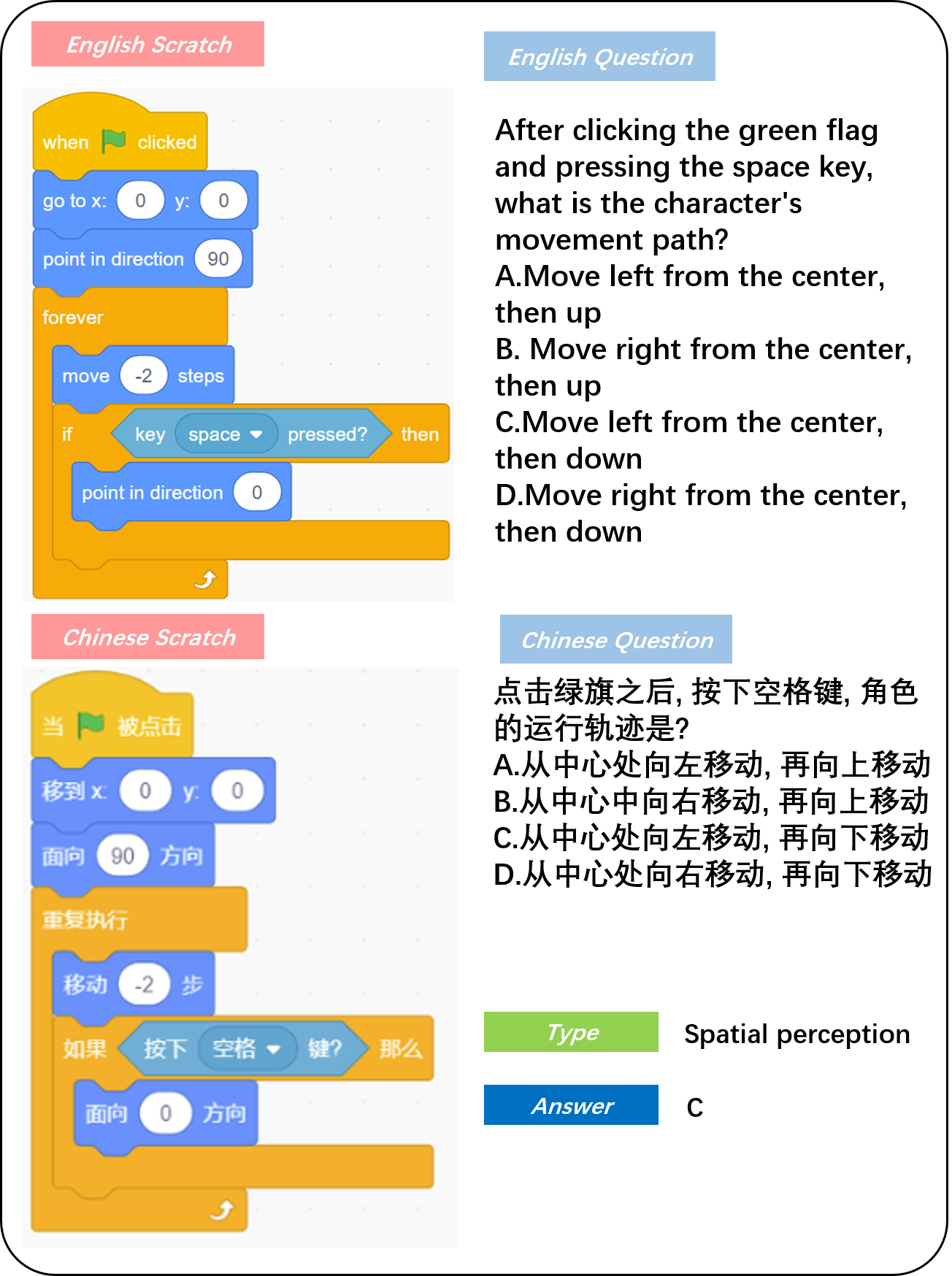}
    \caption{Data example about spatial perception.}
    \label{fig:space}
\end{figure}

In Figure~\ref{fig:math}, Figure~\ref{fig:logic}, Figure~\ref{fig:graphy} and Figure~\ref{fig:space}, we show data for mathematics, logical thinking, graphic perception, and Spatial perception as examples. Each example includes the corresponding Chinese and English scripts, questions, and correct answers.

\subsection{Prompt strategie study data }
\begin{figure}[htbp]
    \centering
    \subfigure[Claude-3.5-Sonnet's performance]{
        \includegraphics[width=0.9\linewidth]{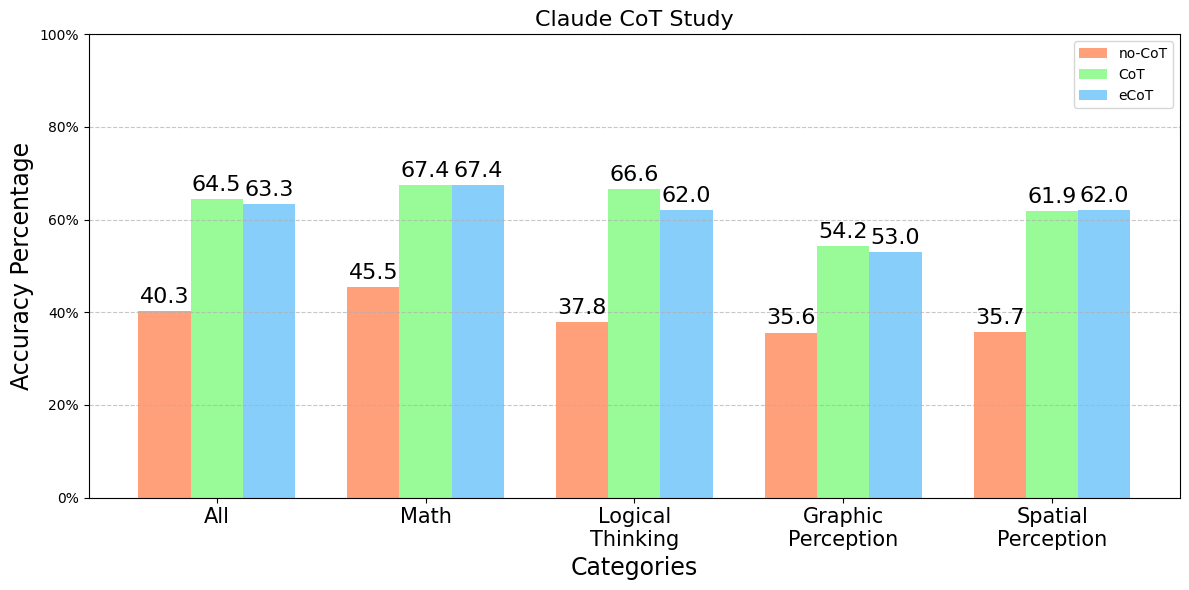}
        \label{fig:claude}
    }
    \subfigure[Gemini-1.5-Pro's performance]{
        \includegraphics[width=0.9\linewidth]{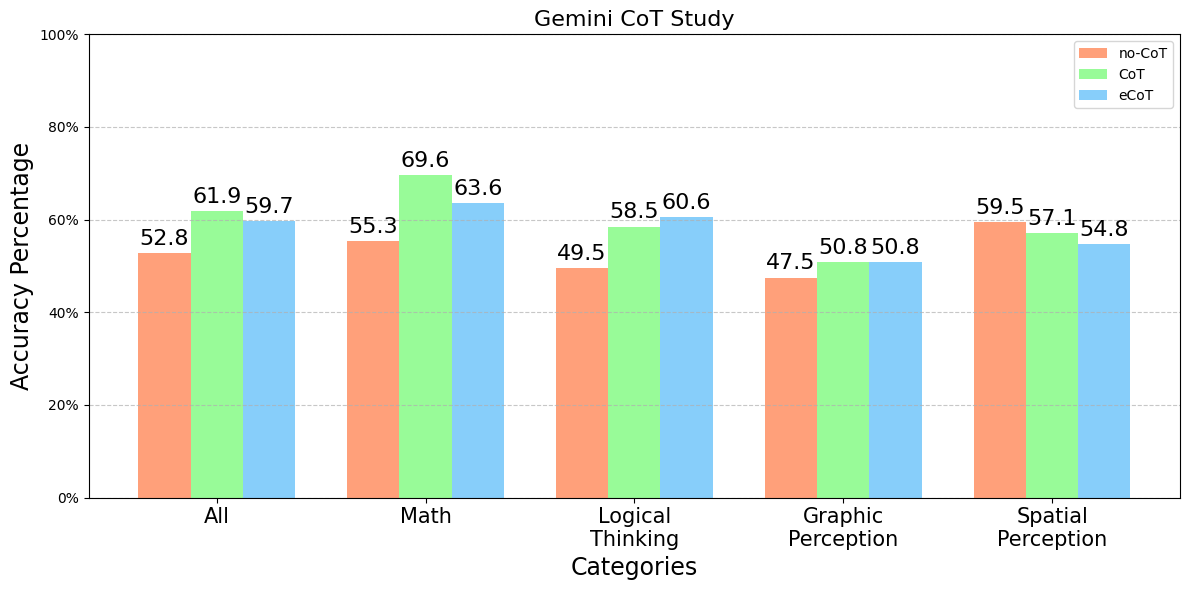}
        \label{fig:gemini}
    }
    \subfigure[GPT-4-Turbo's performance]{
        \includegraphics[width=0.9\linewidth]{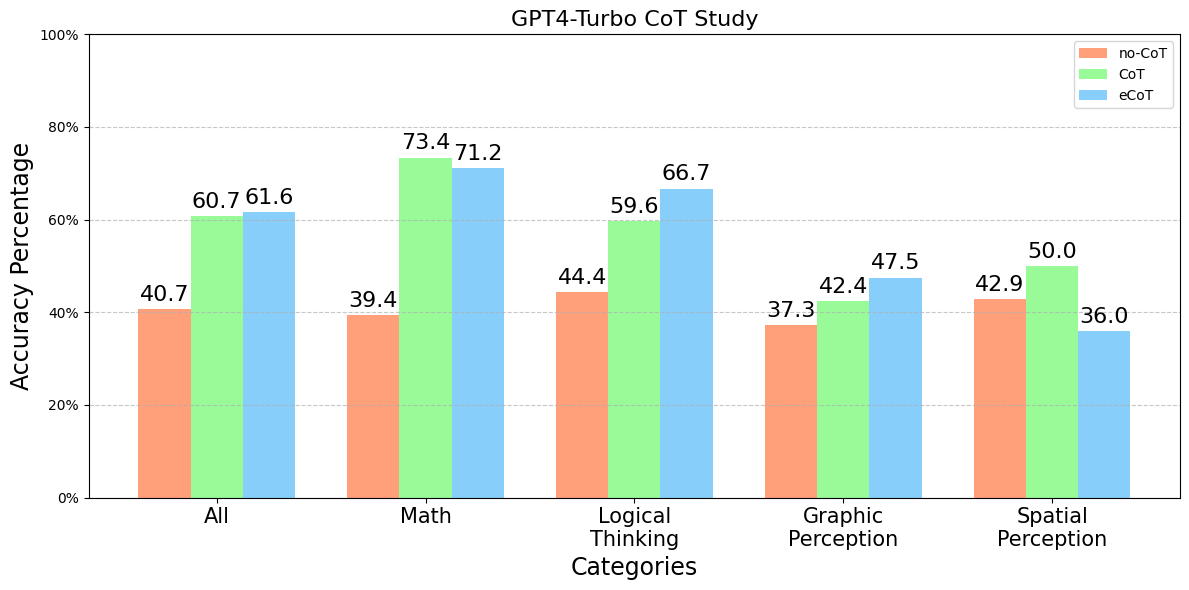}
        \label{fig:gpt4turbo}
    }
    \subfigure[GPT-4o's performance]{
        \includegraphics[width=0.9\linewidth]{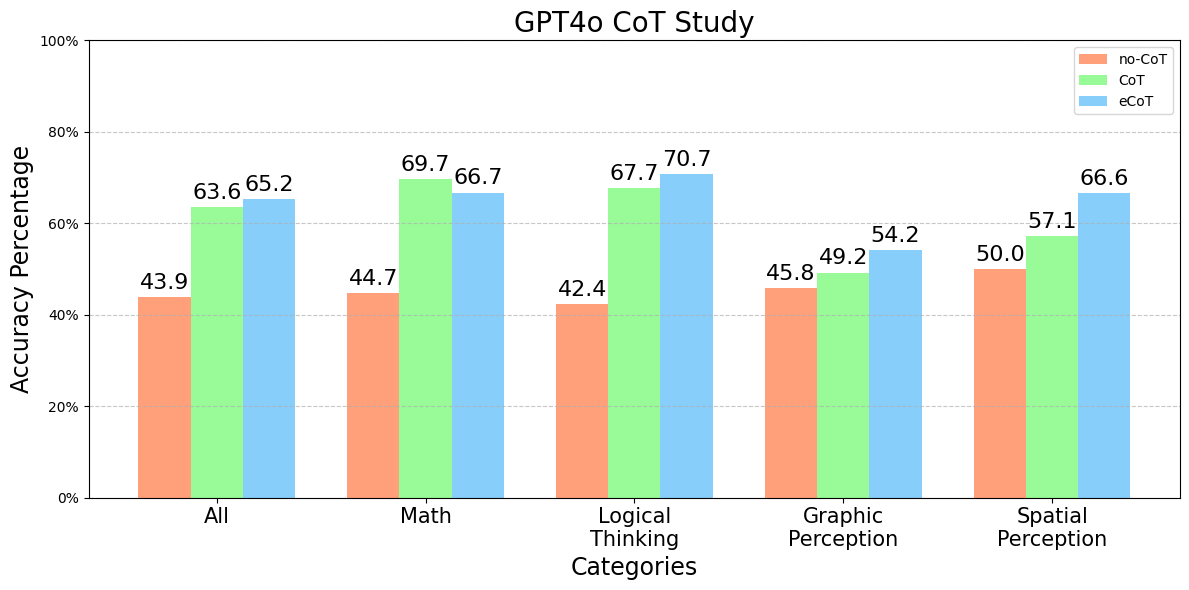}
        \label{fig:gpt4o}
    }
    \caption{Performance under different prompting strategies.}
    \label{fig:all}
\end{figure}

In Figure~\ref{fig:all}, we provide more data on the model performance under different prompt strategies, which are also consistent with the views we put forward in the main text.

\subsection{Examples in case study}
\begin{figure*}[h]
    \centering
    \subfigure[Example 1]{
        \includegraphics[width=0.45\linewidth]{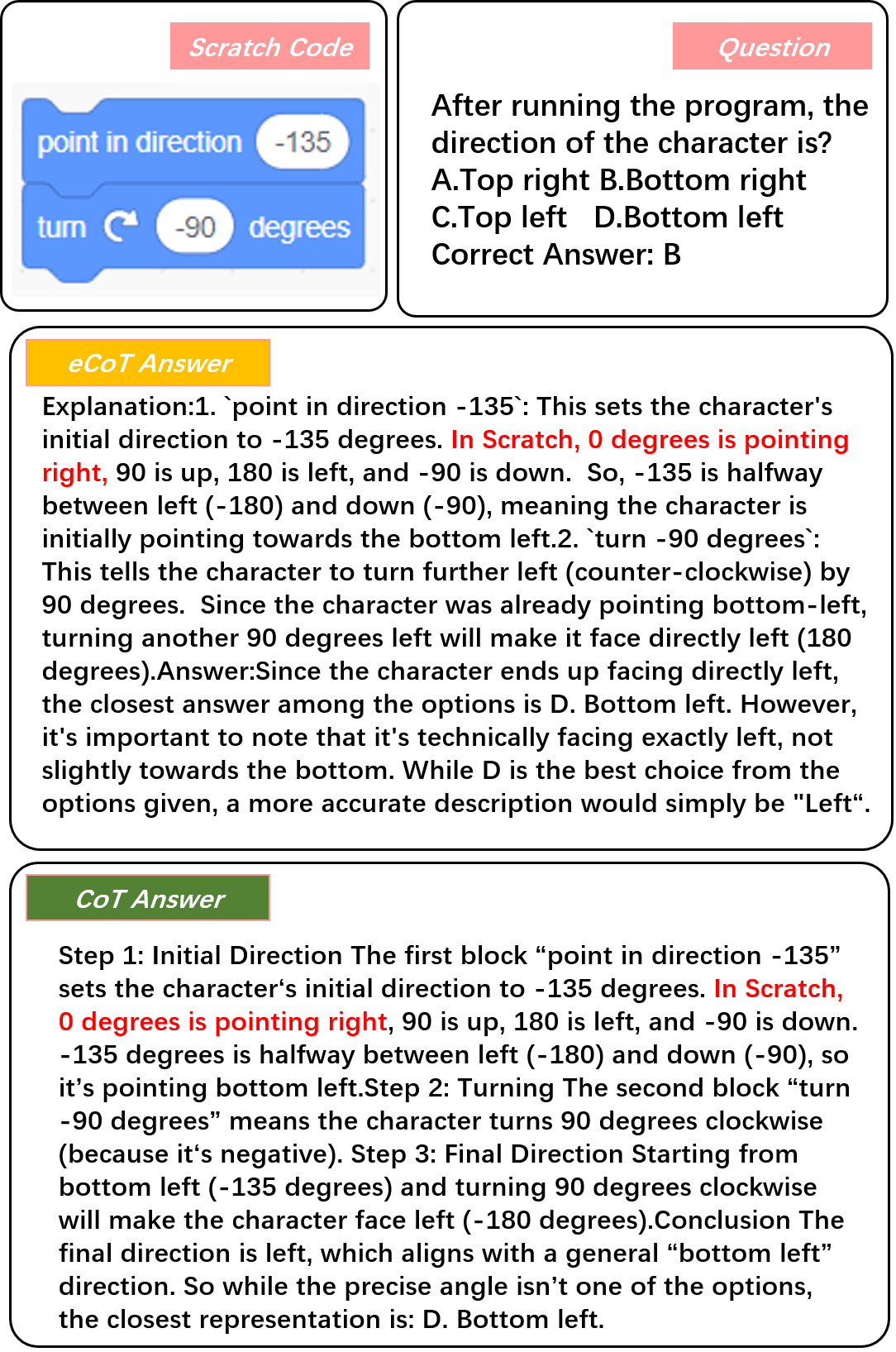}
        \label{fig:eg1}
    }
    \subfigure[Example 2]{
        \includegraphics[width=0.45\linewidth]{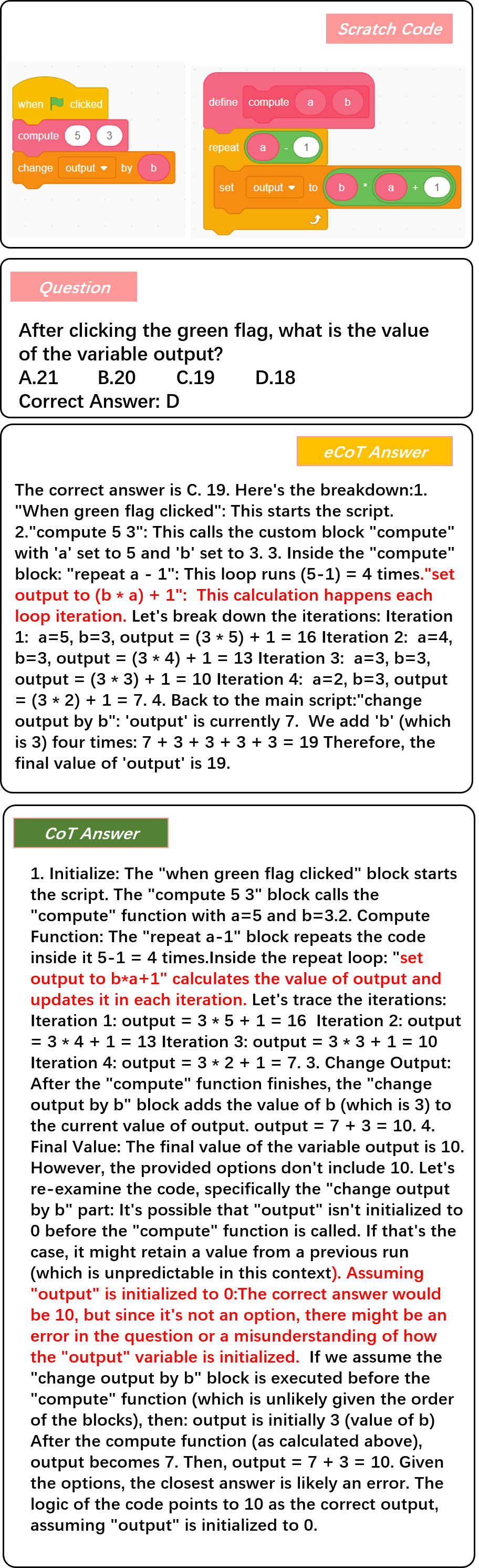}
        \label{fig:eg2}
    }
    \caption{Examples used in the Case study. The error areas are marked in red.}
    \label{fig:case}
\end{figure*}

In Figure~\ref{fig:case}, We show two cases where Gemini-1.5-Pro makes mistakes, and these two cases also illustrate the conclusions we stated in the main text.

\begin{table*}[ht]\small
  \centering
  \renewcommand{\arraystretch}{1.2} 
  \setlength{\tabcolsep}{8pt} 
  \begin{tabular}{lcccccc}
    \toprule
    \textbf{Models} & \textbf{Size} &\textbf{All} & \textbf{Math} & \makecell{\textbf{Logical}\\ \textbf{Thinking}} & \makecell{\textbf{Graphic}\\\textbf{Perception}} & \makecell{\textbf{Spatial}\\\textbf{Perception}}\\
    \midrule
    \rowcolor{pink!50}
    \multicolumn{7}{c}{\textit{Proprietary Models}}\\
    \textbf{Gemini-1.5-Pro}      & -     & \textbf{48.1} & \textbf{52.2} & \textbf{39.4} & \textbf{47.5} & \textbf{54.8} \\
    \textbf{GPT-4o}     & -   & 40.7 & 34.8 & 41.4 & 44.1 & 54.8 \\
    \textbf{GPT-4-Turbo} & -  & 37.4 & 36.4 & 37.4 & 44.1 & 35.7 \\
    \textbf{Claude-3.5-Sonnet}    & -    & 39.7 & 43.2 & 38.9 & 33.9 & 33.3 \\
    
    \midrule
    \rowcolor{green!30}
    \multicolumn{7}{c}{\textit{Open-Source Models}}\\
    \textbf{Qwen2-VL}   & 72B      & \textbf{43.6} & \textbf{43.9} & \textbf{43.4} & \textbf{47.5} & \textbf{40.5} \\
    \textbf{LLaVA-v1.6} & 34B & 28.5 & 20.5 & 34.3 & 33.9 & 31.0 \\
    \textbf{InternVL2} & 26B & 24.3 & 24.2 & 20.2 & 27.1 & 26.2 \\
    \textbf{Pixtral} & 12B & 28.2 & 28.8 & 29.3 & 27.1 & 21.4 \\
    \textbf{MiniCPM-v2.6} & 8B & 30.2 & 28.0 & 29.3 & 37.3 & 26.2 \\
    \textbf{Molmo} & 7B & 30.2 & 28.8 & 30.3 & 33.9 & 26.2 \\
    
    \bottomrule
  \end{tabular}
  \caption{Accuracy (\%) of ten state-of-the-art LMMs on the Chinese data of ScratchEval benchmark, tested across multiple cognitive abilities: math, logical thinking, graphic perception, and spatial perception.}
  \label{tab:cnexperiment_results}
\end{table*}
\label{sec:cndata}

\label{sec:twocase}

\label{sec:allmodelsdata}


\subsection{Potential Risks}

While our benchmark for LMMs, which evaluates models using Scratch visual programming questions, poses no direct risks, potential concerns include the possibility of models overfitting to specific visual programming patterns, reducing their generalization capabilities. Additionally, the reliance on Scratch could limit the applicability of results to broader real-world tasks that use different programming interfaces.

\subsection{Creators Of Artifacts}

The source data for our benchmark is derived from the China Lanqiao Cup National Software and Information Technology Professional Talent Competition \url{https://www.lanqiaoqingshao.cn/home} (Chinese website). To adapt this data for our benchmark, we enlisted the help of domain experts to reannotate and refine the original dataset, ensuring its suitability for evaluating LMMs on Scratch visual programming tasks.

\subsection{License}

The benchmark was annotated and developed by the authors of this paper, and the dataset is released under the Apache 2.0 license.

\subsection{Use Of AI Assistants}

The AI assistant, GPT-4o, was used solely to enhance the writing of this paper.